\documentclass[preprint,12pt]{elsarticle}

\usepackage{amssymb}
\usepackage{epsfig}
\usepackage{epsf}
\usepackage{booktabs}
\usepackage{amssymb}
\usepackage{multirow}
\usepackage{mathrsfs}
\usepackage{longtable}
\usepackage{lscape}
\usepackage{tabularx}
\usepackage{subfigure}
\usepackage{graphicx}
\usepackage{color}

\journal{Knowledge-Based Systems}
\begin{document}

\begin{frontmatter}


\title{A generalized evidence distance}
\author[label1,label5]{Hongming Mo}
\author[label3]{Xiaoyan Su}
\author[label2]{Yong Hu}
\author[label1,label3,label4]{Yong Deng\corref{cor1}}
\ead{ydeng@swu.edu.cn, prof.deng@hotmail.com}
\cortext[cor1]{Corresponding author at: School of Computer and Information Science, Southwest University, Chongqing 400715, China. Tel.: +86 023 68254555; fax: +86 023 68254555. }
\address[label1]{School of Computer and Information Science, Southwest University, Chongqing 400715,China}
\address[label5]{Department of the Tibetan Language, Sichuan University of Nationalities, Kangding  Sichuan 626001, China}
\address[label3]{School of Electronics and Information Technology, Shanghai Jiao Tong University, Shanghai  200240, China}
\address[label2]{Institute of Business Intelligence \& Knowledge Discovery, Department of E-commerce, Guangdong University of Foreign Studies, Sun Yat-Sen University, Guangzhou 510006, China}
\address[label4]{School of Engineering, Vanderbilt University, TN 37235, USA }
\begin{abstract}
 Dempster-Shafer theory of evidence (D-S theory) is widely used in uncertain information process. The basic probability assignment(BPA) is a key element in D-S theory.  How to measure the distance between two BPAs is an open issue. In this paper, a new method to measure the distance of two BPAs is proposed. The proposed method is a generalized of existing evidence distance. Numerical examples are illustrated that the proposed method can overcome the shortcomings of existing methods.
 \end{abstract}

\end{frontmatter}


\section{Introduction}
Dempster-Shafer theory (D-S theory) is first proposed by Dempster, and later by Shafer improved \cite{dempster1968generalization,shafer1976mathematical},  has attracted more and more researchers, for its ability to handle uncertain information in many fields, such as  combination of multi-classier, information fusion, target recognition, decision making, etc\cite{denoeux2008conjunctive,tabassian2012combining,yong2004combining,kang2012evidential}.

Many complex, uncertain information can be well processed by the combination rule of D-S theory. However, when the basic probability assignments (BPAs) are high conflict, the  combination rule of D-S theory will generate an invalid result of counter-intuition.
Besides the coefficient $k$, evidence distance is another expression of measure the conflict of two BPAs. Recently, there are two  distance functions of measure the distance of  two BPAs\cite{jousselme2012distances,jousselme2001new,sunberg2013belief}. One is  proposed by  Jousselme\cite{jousselme2001new}, another is proposed by Sunberg\cite{sunberg2013belief}, based on Hausdorff distance\cite{hausdorff1957set}. The differences between the two methods are the distance functions.
The main point of Jousselme's method is the similarity matrix $\underline{\underline D}$,  measured the conflict of focal elements of BPAs\cite{jousselme2001new}. Jousselme's method is effective  in the case of stationary BPAs. While the BPAs are shifted, it can't reflect  the physical distance of two BPAs, intuitively.
   Sunberg's method is designed specifically for orderable frames of discernment, applied a  Hausdorff-based measure to account for the distance  between focal elements\cite{sunberg2013belief}.  It is suitable for varied BPAs, but can't apply to varied masses. In other words, Jousselme's method is suitable for fixed BPAs but mass changed, Sunberg's method can be applied to  fixed masses  but BPAs changed.
   To address these issues in existing methods,  we propose a generalized  method to measure the evidence distance, 
    that integrates the merits of the existing methods and overcomes the shortcoming of both.

The rest of the this paper is organized as follows. Section 2 presents some preliminaries. The proposed method to measure the distance of two BPAs and numerical examples and applications are presented in Section 3. A short conclusion is drawn in the last section.
\section{Preliminaries}
In D-S theory, Let $\Theta  = \left\{ {\theta _1 ,\theta _2 , \cdots ,\theta _n } \right\}$
be the finite set of mutually exclusive and exhaustive events.
D-S theory is concerned with the set of all subsets of $\Theta$, which is a powerset of $2^{\left| \Theta  \right|}$, known as the frame of discernment, denotes as $\Omega  = \left\{ {\emptyset ,\left\{ {\theta _1 } \right\}, \left\{ {\theta _2 } \right\},\cdots ,\left\{ {\theta _n } \right\},\left\{ {\theta _1 ,\theta _2 } \right\}, \cdots ,\left\{ {\theta _1 ,\theta _1 \cdots ,\theta _n } \right\}} \right\}.$

 The mass function of evidence assigns probability to the subset of $\Omega$, also called basic probability assignment(BPA), which satisfies the following conditions:
$m(\phi ) = 0, 0 \le m(A) \le 1, \sum\limits_{A \subseteq \theta } {m(A) = 1}.$ $\phi$ is an empty set and $A$ is any subsets of $\theta$.

The D-S rule of combination is the first one within the framework of evidence theory which can combine two BPAs $m_1$ and $m_2$ to yield a new BPA $m$. D-S rule of combination are presented as follow:
\begin{equation}
m(A) = \frac{1}{{1 - k}}\sum\limits_{B \cap C = A} {{m_1}(B){m_2}(C)}
\end{equation}

with
\begin{equation}k = \sum\limits_{B \cap C = \emptyset } {m_1 (B)m_2 (C)}\end{equation}

Where  $k$ is a normalization constant, called the conflict coefficient of  BPAs.

There are two existing methods to measure the distance of two BPAs, one is proposed by Jousselme, another  is proposed by Sunberg. The main points of the two methods are shown as follows.

The evidence distance proposed by Jousselme\cite{jousselme2001new}, are presented as follows:
\begin{equation} \label{dpa123}
d_{BPA} (m_1 ,m_2 ) = \sqrt {\frac{1}{2}\left( {{m_1 }  - {m_2 } } \right)^T \underline{\underline D} \left( { {m_1 }  -  {m_2 } } \right)}
\end{equation}

${\underline{\underline D} } $ is an $2^N \times 2^N $ similarity matrix to measure the conflict of focal element in $m_1$ and $m_2 $, where
\begin{equation}\label{DJ}
 \underline{\underline D} (A,B) = \frac{{|A \cap B|}}{{|A \cup B|}}
 \end{equation}


The another evidence distance proposed by Sunberg\cite{sunberg2013belief}, are presented as follows:
\begin{equation}
d_{Haus} (m_1 ,m_2 ) = \sqrt {\frac{1}{2}\left( {{m_1 }  - \ {m_2 } } \right)^T {D_{H}} \left( {{m_1 }  - {m_2 } } \right)}
\end{equation}

with
\begin{equation}\label{Dijhaus}
D_{H(i,j)}=S_{H}(A_{i},A_{j})=\frac{1}{1+KH(A_{i},A_{j})}
\end{equation}

Where H($A_{i}$,$A_{j}$) is the Hausdorff distance between focal elements $A_{i}$ and $A_{j}$. $K>0$ is a user-defined tuning parameter ($K$ is set to be 1, simplified, the same as below). 
It is defined according to
 \begin{equation} \label{DHaus}
{H}(A_{i},A_{j})=\max\{\sup_{b \subseteq A_{i}}\inf_{c \subseteq A_{j}}d(b,c), \sup_{c \subseteq A_{j}}\inf_{b \subseteq A_{i}}d(b,c)\}
\end{equation}

 Where $d(b,c)$ is the distance between two elements of the sets and can be defined as any valid metric distance on the measurement space\cite{hausdorff1957set}.
 In the case where elements of the sets are real numbers, that is, the 1-dimensional Euclidean case, distance can be measured as the absolute value of the difference between the elements\cite{sunberg2013belief}, the Hausdorff distance may be defined as
 \begin{equation}
H_{1-D}(A_{i},A_{j})=\max\{|\min(A_{i})-\min(A_{j})|, |\max(A_{i})-\max(A_{j})|\}
\end{equation}

\section{The generalized evidence distance and Applications}
\subsection{New evidence distance}
Both  Jousselme's and Sunberg's methods can measure the distance of two BPAs, in D-S theory, but the two existing methods take effect only under the special situations, while the cases changed, counter-intuitive results will be presented. In view of these situations, we propose a new method that  synthesize the merits of both Jousselme and Sunberg proposed, suitable for the both special cases, overcome the shortcomings  of the both existing methods.  This new metric proposed mirrors the quadratic form from structure of Jousselme but replaces the distance function. The new metric $d_{New}$ is defined as follows:
\begin{equation}
{d_{New}}({m_1},{m_2}) = \sqrt {\frac{1}{2}{{({m_1} - {m_2})}^T}{D_{New}}({m_1} - {m_2})}
\end{equation}

with
\begin{equation}
{D_{New(i,j)}} = \alpha {\underline{\underline D} _{(i,j)}} + (1 - \alpha ){D_H}_{(i,j)}
\end{equation}

Where $\underline{\underline D} _{(i,j)}$ is the similarity matrix of Jousselme in Eq.(\ref{DJ}), ${D_H}_{(i,j)}$ is the distance matrix of Sunberg in Eq.(\ref{DHaus}). The parameter $\alpha$ is a user-defined weighted normalized coefficient,  $\alpha  \in \left[ {0,1} \right]$. In this paper, $\alpha$ is set to be 0.5.
And the new metric $d_{New}$ satisfies the follow conditions:
\begin{equation}
while \quad \alpha =1,\quad  {d_{New}}({m_1},{m_2})=d_{BPA}({m_1},{m_2})
\end{equation}
\begin{equation}
while  \quad\alpha =0,\quad  {d_{New}}({m_1},{m_2})=d_{Haus}({m_1},{m_2})
\end{equation}

Thus, both the two  methods\cite{jousselme2001new,sunberg2013belief} are special cases of the proposed distance function. The new proposed method of evidence distance is generalized.
\subsection{Numerical examples and Applications} \label{examples}
\subsubsection{Fixed masses of varied BPAs}\label{sunberg}
We use  an example of orderable sets in \cite{sunberg2013belief}.
BPA $A$ is held stationary while the position of BPA $B$ is shifted right along the real line, such that the absolute value of the midpoints of each BPA take on different values (For more detail, please refer to \cite{sunberg2013belief}). In this example, the two BPAs are constructed as follows:

The first BPA $A$ is shown  as
\[\begin{array}{l}
{m_A}(2) = 0.1,\quad {m_A}(2,2.3) = 0.2,\quad{m_A}(2,2.3,2.5) = 0.4,
\\{m_A}(2,2.3,2.5,2.7) = 0.2,\quad\quad{m_A}(2,2.3,2.5,2.7,3) = 0.1
\end{array}\]

The second BPA $B$ is given as
\[{m_B}(i) = \frac{1}{3},\quad{m_B}(i,0.5 + i) = \frac{1}{3},\quad{m_B}(i,0.5 + i,1 + i) = \frac{1}{3}\]

\begin{figure}[!h]
\begin{center}
\psfig{file=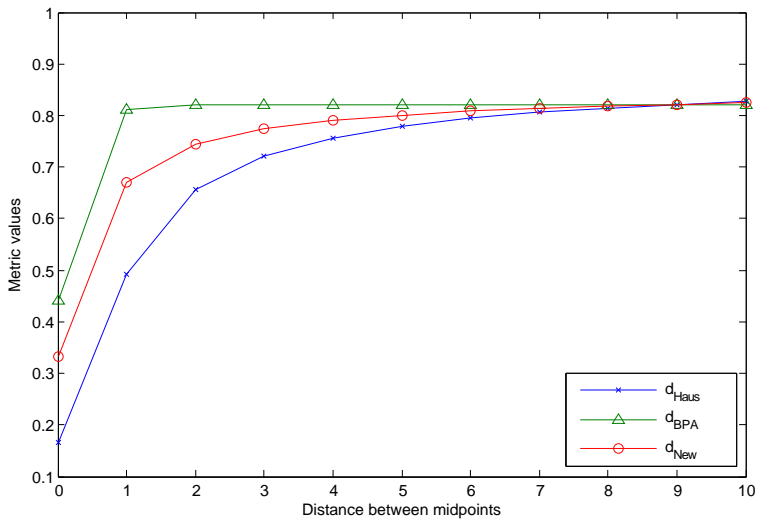,scale=0.6}
\caption{Values for distance metrics as one BPA is held stationary, and another BPA is shifted right to different locations, in the section {\ref{sunberg}}. Note that minimum distances are nonzero due to the difference in structure between the BPAs.}
\label{figsunberg}
\end{center}
\end{figure}

Where $i$ is an integer, varied from 2 to 12, means that $B_{1}$ shifted right along the real line to $B_{2}$.
The distance of the two BPAs are graphically illustrated in the Fig.{\ref{figsunberg}}, through different methods, respectively.
It should be pointed that, while the BPAs are  varied,  Jousselme's method can't reflect the metric of two BPAs effectively, both Sunberg and the new proposed methods can measure the distance of the two BPAs, clearly.
\subsubsection{Varied masses of fixed BPAs} \label{lwr}
 We use an example of varied elements in \cite{liu2006analyzing}. Let $\Omega $ be a frame of discernment with 20 elements (or any number of elements that is pre-defined). The first BPA $m_{1}$  is shown as
\[m_{1}(\{2,3,4\})=0.05, m_{1}(\{7\})=0.05, m_{1}(\Omega)=0.1, m_{1}(A)=0.8 \]

Where $A$ is a subset of  $\Omega$. The second BPA $m_{2}$ is given as
\[ m_{2}(\{1,2,3,4,5\})=1\]

\begin{figure}[!h]
\begin{center}
\psfig{file=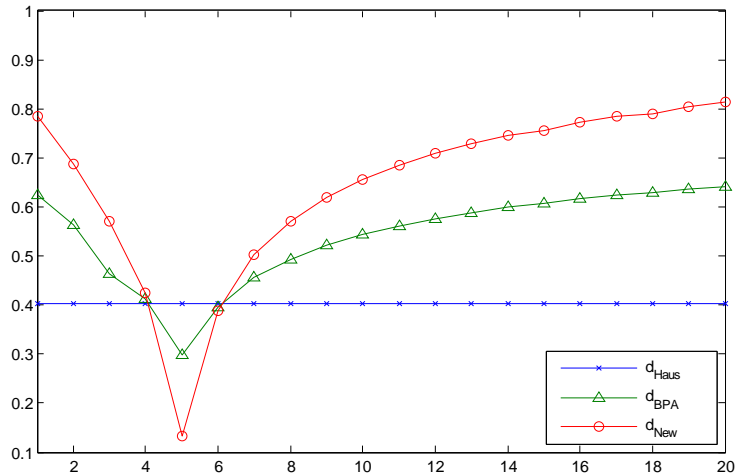,scale=0.6}
\caption{The X-axis shows the sizes of subset $A$, when subset $A$ changes,and the Y-axis show  the different evidence distances, in the section {\ref{lwr}}  used different methods.}
\label{figlwr}
\end{center}
\end{figure}

There are 20 cases where subset $A$ increments one more element at a time, starting from Case 1  with $A$ = \{1\} and ending with Case 20 when $A = \Omega$. The metric of two BPAs are graphically illustrated  in Fig.{\ref{figlwr}}, by the means of  different methods, respectively.
In this example, we notice that the method proposed by Sunberg is inability for the cases of focal elements varied, both Jousselme and our methods can deal with this situation, easily.
\section{Conclusion}
The evidence distance of BPAs plays a key role in D-S theory. In this paper, two  evidence distances are illustrated the shortcomings in some situations. We propose a generalized method to measure the distance of two BPAs. It is generalized and flexible, owing to the adjustable parameter $ \alpha $. Jousselme and Sunberg's methods are the special cases of this proposed generalized method. It integrates the merits and overcomes the shortcomings of existing  method. Numerical examples are demonstrated that the new proposed method can  measure the  distance of two BPAs, effectively.

\section*{Acknowledgement}
 The work is partially supported by National Natural Science Foundation of China (Grant No. 61174022), National High Technology Research and Development Program of China(863 Program)(No. 2013AA013801), Chongqing Natural Science Foundation (Grant No. CSCT, 2010BA2003). 




\bibliographystyle{elsarticle-num}
\bibliography{coresample}

\begin{thebibliography}{10}
\expandafter\ifx\csname url\endcsname\relax
  \def\url#1{\texttt{#1}}\fi
\expandafter\ifx\csname urlprefix\endcsname\relax\def\urlprefix{URL }\fi
\expandafter\ifx\csname href\endcsname\relax
  \def\href#1#2{#2} \def\path#1{#1}\fi

\bibitem{dempster1968generalization}
A.~P. Dempster, A generalization of bayesian inference, Journal of the Royal
  Statistical Society. Series B (Methodological) (1968) 205--247.

\bibitem{shafer1976mathematical}
G.~Shafer, A mathematical theory of evidence, Vol.~1, Princeton university
  press Princeton, 1976.

\bibitem{denoeux2008conjunctive}
T.~Den{\oe}ux, Conjunctive and disjunctive combination of belief functions
  induced by nondistinct bodies of evidence, Artificial Intelligence 172~(2)
  (2008) 234--264.

\bibitem{tabassian2012combining}
M.~Tabassian, R.~Ghaderi, R.~Ebrahimpour, Combining complementary information
  sources in the dempster--shafer framework for solving classification problems
  with imperfect labels, Knowledge-Based Systems 27 (2012) 92--102.

\bibitem{yong2004combining}
Y.~Deng, S.~WenKang, Z.~ZhenFu, L.~Qi, Combining belief functions based on
  distance of evidence, Decision Support Systems 38~(3) (2004) 489--493.

\bibitem{kang2012evidential}
B.~Kang, Y.~Deng, R.~Sadiq, S.~Mahadevan, Evidential cognitive maps,
  Knowledge-Based Systems 35 (2012) 77--86.

\bibitem{jousselme2012distances}
A.-L. Jousselme, P.~Maupin, Distances in evidence theory: Comprehensive survey
  and generalizations, International Journal of Approximate Reasoning 53~(2)
  (2012) 118--145.

\bibitem{jousselme2001new}
A.-L. Jousselme, D.~Grenier, {\'E}.~Boss{\'e}, A new distance between two
  bodies of evidence, Information fusion 2~(2) (2001) 91--101.

\bibitem{sunberg2013belief}
Z.~Sunberg, J.~Rogers, A belief function distance metric for orderable sets,
  Information Fusion 14~(4) (2013) 361--373.

\bibitem{hausdorff1957set}
F.~Hausdorff, Set Theory: Translated from the German by John R. Aumann, Et Al,
  Vol. 119, AMS Bookstore, 1957.

\bibitem{liu2006analyzing}
W.~Liu, Analyzing the degree of conflict among belief functions, Artificial
  Intelligence 170~(11) (2006) 909--924.

\end{thebibliography}







\end{document}